\documentclass[11pt,letterpaper]{article}
\usepackage{emnlp2017}
\usepackage{times}
\usepackage{latexsym}

\usepackage{graphicx}
\usepackage{amsmath}
\usepackage{soul}
\usepackage{bm}
\usepackage{bbm}
\usepackage{caption}

\emnlpfinalcopy

\def\emnlppaperid{836}

\title{CRF Autoencoder for Unsupervised Dependency Parsing\Thanks{This work was supported by the National Natural Science Foundation of China (61503248).}}

\author{Jiong Cai, Yong Jiang \and Kewei Tu \\
  {\tt \{caijiong,jiangyong,  tukw\}@shanghaitech.edu.cn}\\
  School of Information Science and Technology\\ShanghaiTech University, Shanghai, China
}

\date{}

\begin{document}

\maketitle

\begin{abstract}
Unsupervised dependency parsing, which tries to discover linguistic dependency structures from unannotated data, is a very challenging task. Almost all previous work on this task focuses on learning generative models. In this paper, we develop an unsupervised dependency parsing model based on the CRF autoencoder. The encoder part of our model is discriminative and globally normalized which allows us to use rich features as well as universal linguistic priors. We propose an exact algorithm for parsing as well as a tractable learning algorithm. We evaluated the performance of our model on eight multilingual treebanks and found that our model achieved comparable performance with state-of-the-art approaches.
\end{abstract}


\section{Introduction}
Unsupervised dependency parsing, which aims to discover syntactic structures in sentences from unlabeled data, is a very challenging task in natural language processing. Most of the previous work on unsupervised dependency parsing is based on generative models such as the dependency model with valence (DMV) introduced by Klein and Manning \shortcite{klein2004corpus}. Many approaches have been proposed to enhance these generative models, for example, by designing advanced Bayesian priors \cite{cohen2008logistic}, representing dependencies with features \cite{berg2010painless}, and representing discrete tokens with continuous vectors \cite{jiang-han-tu:2016:EMNLP2016}.

Besides generative approaches, Grave and Elhadad \shortcite{grave2015convex} proposed an unsupervised discriminative parser. They designed a convex quadratic objective function under the discriminative clustering framework. By utilizing global features and linguistic priors, their approach achieves state-of-the-art performance. However, their approach uses an approximate parsing algorithm, which has no theoretical guarantee. In addition, the performance of the approach depends on a set of manually specified linguistic priors.

Conditional random field autoencoder \cite{ammar2014conditional} is a new framework for unsupervised structured prediction. There are two components of this model: an encoder and a decoder. The encoder is a globally normalized feature-rich CRF model predicting the conditional distribution of the latent structure given the observed structured input.  The decoder of the model is a generative model generating a transformation of the structured input from the latent structure. Ammar et al. \shortcite{ammar2014conditional} applied the model to two sequential structured prediction tasks, part-of-speech induction and word alignment and showed that by utilizing context information the model can achieve better performance than previous generative models and locally normalized models. However, to the best of our knowledge, there is no previous work applying the CRF autoencoder to tasks with more complicated outputs such as tree structures.

In this paper, we propose an unsupervised discriminative dependency parser based on the CRF autoencoder framework and provide tractable algorithms for learning and parsing. We performed experiments in eight languages and show that our approach achieves comparable results with previous state-of-the-art models.

\section{Method}

\subsection{Model}
\label{sec:model}
\begin{figure}[t]
    \centering
    \includegraphics[width=\columnwidth]{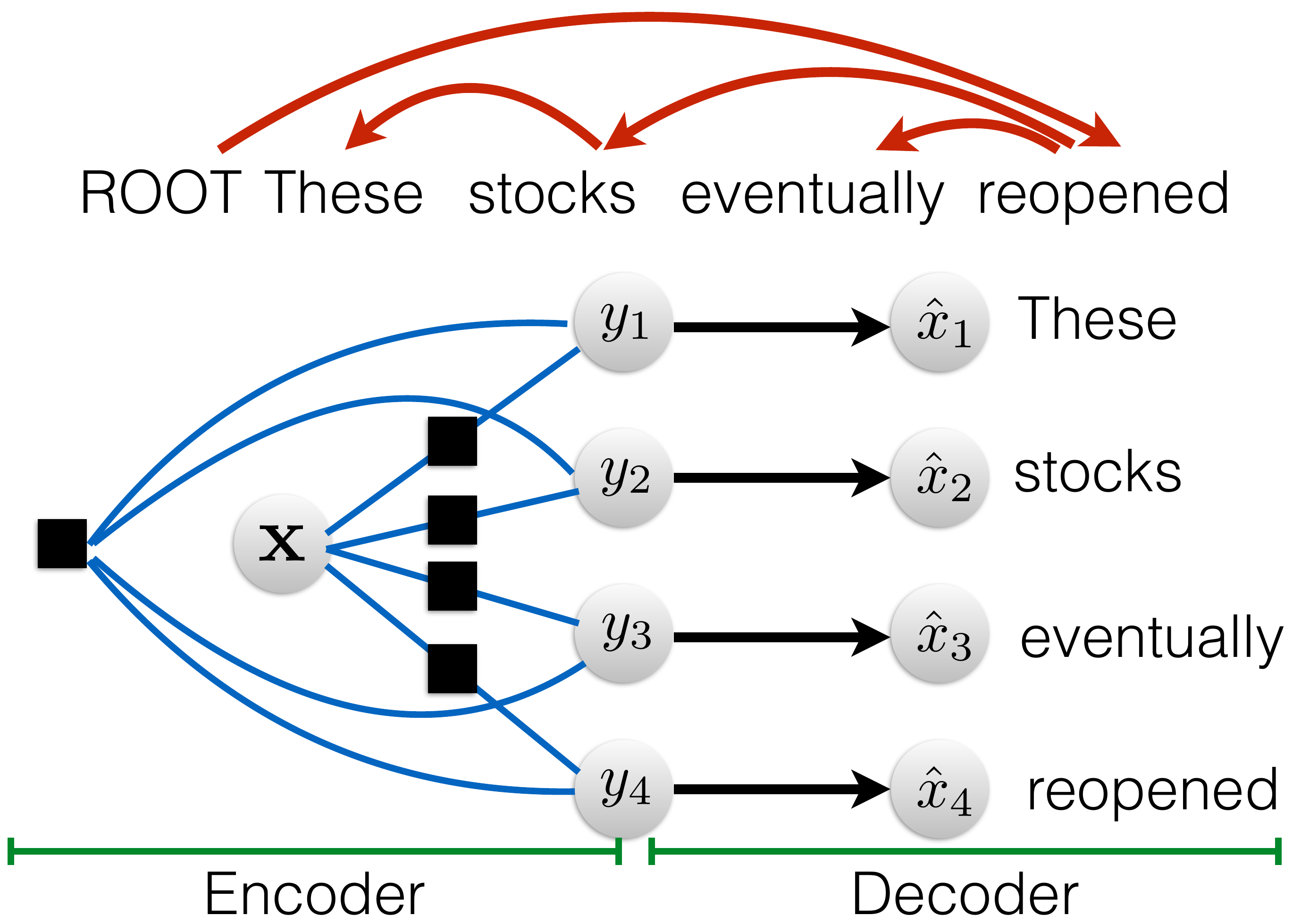}
\caption{The CRF Autoencoder for the input sentence ``These stocks eventually reopened'' and its corresponding parse tree (shown at the top). $\mathbf{x}$ and $\hat{\mathbf{x}}$ are the original and reconstructed sentence. $\mathbf{y}$ is the dependency parse tree represented by a sequence where $y_i$ contains the token and index of the parent of token $x_i$ in the parse tree, e.g., $y_1=\langle \text{stocks}, 2\rangle$ and $y_2=\langle \text{reopened},4\rangle.$ 
The encoder is represented by a factor graph (with a global factor specifying valid parse trees) and the decoder is represented by a Bayesian net.}
    \label{fig:example}
\end{figure}

Figure \ref{fig:example} shows our model with an example input sentence.
Given an input sentence $\mathbf{x} = (x_1, x_2, \ldots, x_n)$, we regard its parse tree as the latent structure represented by a sequence $\mathbf{y} = (y_1, y_2, \ldots, y_n)$ where $y_i$ is a pair $\langle t_i, h_i \rangle$, $t_i$ is the head token of the dependency connecting to token $x_i$ in the parse tree, and $h_i$ is the index of this head token in the sentence.
The model also contains a reconstruction output, which is a token sequence $\hat{\mathbf{x}} = (\hat{x}_1, \hat{x}_2, \ldots, \hat{x}_n)$. Throughout this paper, we set $\hat{\mathbf{x}} = \mathbf{x}$.

The encoder in our model is a log-linear model represented by a first-order dependency parser. The score of a dependency tree can be factorized as the sum of scores of its dependencies. For each dependency arc $(\mathbf{x}, i, j)$, where $i$ and $j$ are the indices of the head and child of the dependency, a feature vector $\mathbf{f}(\mathbf{x}, i, j)$ is specified. The score of a dependency is defined as the inner product of the feature vector and a weight vector $\mathbf{w}$,
\[
\phi(\mathbf{x}, i, j) = \mathbf{w}^T \mathbf{f}(\mathbf{x}, i, j)
\]
The score of a dependency tree $\mathbf{y}$ of sentence $\mathbf{x}$ is 
\[
    \phi(\mathbf{x}, \mathbf{y}) = \sum_{i=1}^n {\phi(\mathbf{x}, h_i, i)}
\]
We define the probability of parse tree $\mathbf{y}$ given sentence $\mathbf{x}$ as 
\[
    P(\mathbf{y}|\mathbf{x}) = \frac{\exp(\phi(\mathbf{x}, \mathbf{y}))}{Z(\mathbf{x})}
\]
$Z(\mathbf{x})$ is the partition function,
\[Z(\mathbf{x}) = \sum_{\mathbf{y}' \in \mathcal{Y}(\mathbf{x})}{\exp(\phi(\mathbf{x}, \mathbf{y}'))}\]
where $\mathcal{Y}(\mathbf{x})$ is the set of all valid parse trees of $\mathbf{x}$. The partition function can be efficiently computed in $O(n^3)$ time using a variant of the inside-outside algorithm \cite{paskin2001cubic} for projective tree structures, or using the Matrix-Tree Theorem for non-projective tree structures \cite{koo2007structured}.

The decoder of our model consists of a set of categorical conditional distributions $\theta_{x|t}$, which represents the probability of generating token $x$ conditioned on token $t$. So the probability of the reconstruction output $\hat{\mathbf{x}}$ given the parse tree $\mathbf{y}$ is
\[
    P(\hat{\mathbf{x}}|\mathbf{y}) = \prod_{i=1}^n {\theta_{\hat{x}_i | t_i}} 
\]

The conditional distribution of $\hat{\mathbf{x}}, \mathbf{y}$ given $\mathbf{x}$ is
\[
    P(\mathbf{y}, \hat{\mathbf{x}}|\mathbf{x}) = P(\mathbf{y} | \mathbf{x}) P(\hat{\mathbf{x}}|\mathbf{y})
\]

\subsubsection{Features}
Following McDonald et al. \shortcite{mcdonald2005online} and Grave et al. \shortcite{grave2015convex}, we define the feature vector of a dependency based on the part-of-speech tags (POS) of the head, child and context words, the direction, and the distance between the head and child of the dependency. 
The feature template used in our parser is shown in Table \ref{tab:feat}.

\begin{table}
\centering
\begin{tabular}{|l|}
\hline
$\text{POS}_i \times dis \times dir$ \\
$\text{POS}_j \times dis \times dir$ \\
\hline
$\text{POS}_i \times \text{POS}_j \times dis \times dir$ \\
\hline
$\text{POS}_i \times \text{POS}_{i-1} \times \text{POS}_j \times dis \times dir$ \\
$\text{POS}_i \times \text{POS}_{i+1} \times \text{POS}_j \times dis \times dir$ \\
$\text{POS}_i \times \text{POS}_j \times \text{POS}_{j-1} \times dis \times dir$ \\
$\text{POS}_i \times \text{POS}_j \times \text{POS}_{j+1} \times dis \times dir$ \\
\hline
\end{tabular}
\caption{Feature template of a dependency, where $i$ is the index of the head, $j$ is the index of the child, $dis = |i - j|$, and $dir$ is the direction of the dependency.}\label{tab:feat}
\end{table}

\subsubsection{Parsing}
Given parameters $\mathbf{w}$ and $\bm{\theta}$, we can parse a sentence $\mathbf{x}$ by searching for a dependency tree $\mathbf{y}$ which has the highest probability $P(\hat{\mathbf{x}}, \mathbf{y} | \mathbf{x})$.

\[
\begin{split}
      \mathbf{y}^* & = \arg \max_{\mathbf{y} \in \mathcal{Y}(\mathbf{x})} {\log P(\hat{\mathbf{x}}, \mathbf{y} | \mathbf{x})} \\
      & = \arg \max_{\mathbf{y} \in \mathcal{Y}(\mathbf{x})} {\sum_{i=1}^n \left({\phi(\mathbf{x}, h_i, i) + \log{\theta_{\hat{x}_i | t_i}}}\right)} 
\end{split}
\]

For projective dependency parsing, we can use Eisner’s algorithm \shortcite{eisner1996three} to find the best parse in $O(n^3)$ time. For non-projective dependency parsing, we can use the Chu-Liu/Edmond algorithm \cite{chu1965shortest,edmonds1967optimum,tarjan1977finding} to find the best parse in $O(n^2)$ time.

\subsection{Parameter Learning}
\label{sec:learning}

\subsubsection{Objective Function}
Spitkovsky et al. \shortcite{spitkovsky2010viterbi} shows that Viterbi EM can improve the performance of unsupervised dependency parsing in comparison with EM. Therefore, instead of using negative conditional log likelihood as our objective function, we choose to use negative conditional Viterbi log likelihood,
\begin{align}\label{obj}
-\sum_{i=1}^{N}\log\left(\max_{\mathbf{y} \in \mathcal{Y}(\mathbf{x}_i)}{P(\hat{\mathbf{x}}_i, \mathbf{y} | \mathbf{x}_i)}\right) + \lambda \Omega(\mathbf{w})
\end{align}
where $\Omega(\mathbf{w})$ is a L1 regularization term of the encoder parameter $\mathbf{w}$ and $\lambda$ is a hyper-parameter controlling the strength of regularization.

To encourage learning of dependency relations that satisfy universal linguistic knowledge, we add a soft constraint on the parse tree based on the universal syntactic rules following Naseem et al.  \shortcite{naseem2010using} and Grave et al. \shortcite{grave2015convex}. Hence our objective function becomes 
\begin{align*}
 -\sum_{i=1}^{N} \log\left(\max_{\mathbf{y}\in \mathcal{Y}(\mathbf{x}_i)}\!{P(\hat{\mathbf{x}}_i, \mathbf{y} | \mathbf{x}_i)  Q^\alpha(\mathbf{x}_i, \mathbf{y})}\right)\!+\!\lambda \Omega(\mathbf{w})   
\end{align*}
where $Q(\mathbf{x}, \mathbf{y})$ is a soft constraint factor over the parse tree, and $\alpha$ is a hyper-parameter controlling the strength of the constraint factor. The factor $Q$ is also decomposable by edges in the same way as the encoder and the decoder, and therefore our parsing algorithm can still be used with this factor taken into account.
\[
Q(\mathbf{x}, \mathbf{y}) = \exp\left({\sum_i {\mathbbm{1}[(t_i \rightarrow x_i) \in \mathcal{R}] }}\right)
\]
where $\mathbbm{1}[( t_i \rightarrow x_i) \in \mathcal{R}]$ is an indicator function of whether dependency $ t_i \rightarrow x_i $ satisfies one of the universal linguistic rules in $\mathcal{R}$. The universal linguistic rules that we use are shown in Table~\ref{tab:rules} \cite{naseem2010using}.

\begin{table}
\centering
\begin{tabular}{|c | c |}
\hline
VERB $\rightarrow$ VERB & NOUN $\rightarrow$ NOUN \\
\hline
VERB $\rightarrow$ NOUN & NOUN $\rightarrow$ ADJ \\
\hline
VERB $\rightarrow$ PRON & NOUN $\rightarrow$ DET \\
\hline
VERB $\rightarrow$ ADV &  NOUN $\rightarrow$ NUM \\
\hline
VERB $\rightarrow$ ADP &  NOUN $\rightarrow$ CONJ \\
\hline
ADJ $\rightarrow$ ADV & ADP $\rightarrow$ NOUN\\
\hline
\end{tabular}
\caption{Universal linguistic rules} \label{tab:rules}
\end{table}

\begin{table*}[t]
\setcounter{table}{3}
\small
\centering
\begin{tabular}{|l|c|c|c|c|c|c|c|c|}
\hline
                           & Basque & Czech & Danish & Dutch & Portuguese & Slovene & Swedish & Avg \\ \hline
\multicolumn{9}{|c|}{Length: $\leq$ 10} \\ \hline 
DMV(EM)                 &41.1    &31.3    &50.8    &47.1    &36.7    &36.7    &43.5    &41.0\\ \hline
DMV(Viterbi)            &47.1    &27.1    &39.1    &37.1    &32.3    &23.7    &42.6    &35.5 \\  \hline
Neural DMV (EM)    &46.5    &33.1    &\bf{55.6}    &\bf{49.0}    &30.4    &42.2    &44.3    &43.0 \\\hline
Neural DMV (Viterbi)    &48.1    &28.6    &39.8    &37.2    &36.5    &39.9    &47.9    &39.7 \\\hline
Convex-MST (No Prior)    &29.4   &36.5    &49.3   &31.3   &46.4   &33.7    &35.5   &37.4\\ \hline
Convex-MST (With Prior)          &30.0    &46.1    &51.6    &35.3    &55.4    &\bf{63.7}    &50.9    &47.5\\ \hline
CRFAE (No Prior)          &49.0   &33.9    &28.8     &39.3     &47.6     &34.7    &51.3    &40.6 \\ \hline
CRFAE (With Prior)              &\bf{49.9}    &\bf{48.1}    &53.4    &43.9    &\bf{68.0}    &52.5    &\bf{64.7}    &\bf{54.3}\\ \hline

\multicolumn{9}{|c|}{Length: All} \\ \hline 
DMV(EM)                 &31.2    &28.1    &40.3    &44.2    &23.5    &25.2    &32.0    &32.0 \\ \hline
DMV(Viterbi)            &40.9    &20.4    &32.6    &33.0    &26.9    &16.5    &36.2    &29.5 \\ \hline
Neural DMV (EM)    &38.5    &29.3    &\bf{46.1}    &\bf{46.2}    &16.2    &36.6    &32.8    &35.1 \\\hline
Neural DMV (Viterbi)    &\bf{41.8}    &23.8    &34.2    &33.6    &29.4    &30.8    &40.2    &33.4\\ \hline        
Convex-MST (No Prior)      &30.5   &33.4   &44.2   &29.3   &38.3   &32.2   &28.3   &33.7\\ \hline
Convex-MST (With Prior)       &30.6    &\bf{40.0}    &45.8    &35.6    &46.3    &\bf{51.8}    &40.5    &41.5\\\hline
CRFAE (No Prior)          &39.8   &25.4   &24.2   &35.2   &52.2   &26.4   &40.0   &34.7\\ \hline
CRFAE (With Prior)              &41.4    &36.8    &40.5    &38.6    &\bf{58.9}    &43.3    &\bf{48.5}    &\bf{44.0}\\ \hline

\end{tabular}
\caption{Parsing accuracy on seven languages. Our model is compared with DMV \cite{klein2004corpus}, Neural DMV \cite{jiang-han-tu:2016:EMNLP2016}, and Convex-MST \cite{grave2015convex}} \label{tab:pascal}
\end{table*}

\begin{table}[t]\centering
\setcounter{table}{2}
\small
\begin{tabular}{|l|l|l|}
\hline
{\bf Methods} & {\bf  WSJ10} & {\bf WSJ}\\\hline
\multicolumn{3}{|c|}{Basic Setup} \\ \hline
Feature DMV \tiny  \cite{berg2010painless} & 63.0 &  ~~~- \\
UR-A E-DMV \tiny \cite{tu2012unambiguity} & 71.4 & 57.0\\ 
Neural E-DMV \tiny \cite{jiang-han-tu:2016:EMNLP2016} & 69.7 & 52.5\\
Neural E-DMV \tiny (Good Init) \tiny \cite{jiang-han-tu:2016:EMNLP2016} & 72.5 & 57.6\\
\hline
\multicolumn{3}{|c|}{Basic Setup + Universal Linguistic Prior} \\ \hline
Convex-MST \tiny \cite{grave2015convex} & 60.8 & 48.6 \\
HDP-DEP \tiny \cite{naseem2010using}& 71.9 & ~~~- \\
CRFAE & 71.7 & 55.7 \\
\hline
\multicolumn{3}{|c|}{Systems Using Extra Info} \\ \hline
LexTSG-DMV \tiny \cite{blunsom2010unsupervised} & 67.7 & 55.7 \\ 
CS \tiny \cite{spitkovsky2013breaking} & 72.0 & 64.4\\
MaxEnc \tiny \cite{le2015unsupervised} & 73.2 & 65.8\\ 
\hline
\end{tabular}
\caption{Comparison of recent unsupervised dependency parsing systems on English. Basic setup is the same as our setup except that linguistic prior is not used. Extra info includes lexicalization, longer training sentences, etc.}
\label{tab:result_eng}
\end{table}

\subsubsection{Algorithm}
We apply coordinate descent to minimize the objective function, which alternately updates $\mathbf{w}$ and $\bm{\theta}$. In each optimization step of $\mathbf{w}$, we run two epochs of stochastic gradient descent, and in each optimization step of $\bm{\theta}$, we run two iterations of the Viterbi EM algorithm.

To update $\mathbf{w}$ using stochastic gradient descent, for each sentence $\mathbf{x}$, we first run the parsing algorithm to find the best parse tree $\mathbf{y^*} = \arg\max_{\mathbf{y}\in \mathcal{Y(\mathbf{x})}}(P(\hat{\mathbf{x}}, \mathbf{y}| \mathbf{x}) Q^{\alpha} (\mathbf{x},\mathbf{y}))$; then we can calculate the gradient of the objective function based on the following derivation,
\[
\begin{split}
    & \frac{\partial{\log{P(\hat{\mathbf{x}}, \mathbf{y}^* | \mathbf{x})}}}{\partial{\mathbf{w}}}  \\
    & = \frac{\partial{\log P(\mathbf{y^*}|\mathbf{x})}}{\partial{\mathbf{w}}} + \frac{\partial{\log P(\hat{\mathbf{x}}|\mathbf{y^*})}}{\partial{\mathbf{w}}}  \\
    & = \frac{\partial{\log P(\mathbf{y^*}|\mathbf{x})}}{\partial{\mathbf{w}}} \\
    & = \frac{\partial{ \bigl(\sum_{i=1}^n{\mathbf{w}^T\mathbf{f}(\mathbf{x}, h_i, i)} - \log Z(\mathbf{x})} \bigr) }{\partial{\mathbf{w}}} \\
    & = \sum_{(i, j) \in \mathbf{D(x)}} f(\mathbf{x}, i, j) \biggl( \mathbbm{1}[ y^*_j = \langle i, x_i \rangle]
    - \mu(\mathbf{x}, i, j) \biggr)
\end{split}
\]
where $\mathbf{D(x)}$ is the set of all possible dependency arcs of sentence $\mathbf{x}$, $\mathbbm{1}[\cdot]$ is the indicator function, and $\mu(\mathbf{x}, i, j)$ is the expected count defined as follows,
\[
\mu(\mathbf{x}, i, j) = \sum_{\mathbf{y} \in \mathcal{Y}(\mathbf{x})}{\bigl( \mathbbm{1}[ y_j = \langle i, x_i \rangle ]P(\mathbf{y} | \mathbf{x}) \bigr)}
\]
The expected count can be efficiently computed using the Matrix-Tree Theorem \cite{koo2007structured} for non-projective tree structures or using a variant of the inside-outside algorithm for projective tree structures \cite{paskin2001cubic}.

To update $\bm{\theta}$ using Viterbi EM, in the E-step we again use the parsing algorithm to find the best parse tree $\mathbf{y}^*$ for each sentence $\mathbf{x}$; then in the M-step we update $\bm{\theta}$ by maximum likelihood estimation.
\[
    \theta_{c | t} = 
    \frac{
        \sum_\mathbf{x} \sum_i { \mathbbm{1}[ x_i=c, y^*_i=\langle \cdot, t \rangle ] }
    }{
        \sum_{c^{\prime}}{\sum_\mathbf{x} \sum_i { \mathbbm{1}[ x_i= c^{\prime}, y^*_i=\langle \cdot, t \rangle ] }}
    }
\]

\section{Experiments}
\subsection{Setup}
We experimented with projective parsing and used the informed initialization method proposed by Klein and Manning \shortcite{klein2004corpus} to initialize our model before learning. We tested our model both with and without using the universal linguistic rules. We used AdaGrad to optimize $\mathbf{w}$. We used POS tags of the input sentence as the tokens in our model. We learned our model on training sentences of length $\leq$ 10 and reported the directed dependency accuracy on test sentences of length $\leq$ 10 and on all test sentences.

\subsection{Results on English}
We evaluated our model on the Wall Street Journal corpus. We trained our model on section 2-21, tuned the hyperparameters on section 22, and tested our model on section 23. Table \ref{tab:result_eng} shows the directed dependency accuracy of our model (CRFAE) compared with recently published results. It can be seen that our method achieves a comparable performance with state-of-the-art systems. 

We also compared the performances of CRF autoencoder using an objective function with negative log likelihood vs. using our Viterbi version of the objective function (Eq.\ref{obj}). We find that the Viterbi version leads to much better performance (55.7 vs. 41.8 in parsing accuracy of WSJ), which echoes Spitkovsky et al. 's findings on Viterbi EM \shortcite{spitkovsky2010viterbi}.

\subsection{Multilingual Results}
We evaluated our model on seven languages from the PASCAL Challenge on Grammar Induction \cite{gelling2012pascal}. We did not use the Arabic corpus because the number of training sentences with length $\leq$ 10 is less than 1000. 
The result is shown in Table~\ref{tab:pascal}.
The accuracies of DMV and Neural DMV are from Jiang et.al \shortcite{jiang-han-tu:2016:EMNLP2016}.
Both our model (CRFAE) and Convex-MST were tuned on the validation set of each corpus. It can be seen that our method achieves the best results on average. Besides, our method outperforms Convex-MST both with and without linguistic prior. From the results we can also see that utilizing universal linguistic prior greatly improves the performance of Convex-MST and our model.

\section{Conclusion}
In this paper, we propose a new discriminative model for unsupervised dependency parsing based on  CRF autoencoder. Both learning and inference of our model are tractable. We tested our method on eight languages and show that our model is competitive to the state-of-the-art systems. 

The code is available at \url{https://github.com/caijiong/CRFAE-Dep-Parser}

\section*{}

\bibliography{emnlp2017}

\begin{thebibliography}{20}
\expandafter\ifx\csname natexlab\endcsname\relax\def\natexlab#1{#1}\fi

\bibitem[{Ammar et~al.(2014)Ammar, Dyer, and Smith}]{ammar2014conditional}
Waleed Ammar, Chris Dyer, and Noah~A Smith. 2014.
\newblock Conditional random field autoencoders for unsupervised structured
  prediction.
\newblock In \emph{Advances in Neural Information Processing Systems}, pages
  3311--3319.

\bibitem[{Berg-Kirkpatrick et~al.(2010)Berg-Kirkpatrick, Bouchard-C{\^o}t{\'e},
  DeNero, and Klein}]{berg2010painless}
Taylor Berg-Kirkpatrick, Alexandre Bouchard-C{\^o}t{\'e}, John DeNero, and Dan
  Klein. 2010.
\newblock Painless unsupervised learning with features.
\newblock In \emph{Human Language Technologies: The 2010 Annual Conference of
  the North American Chapter of the Association for Computational Linguistics},
  pages 582--590. Association for Computational Linguistics.

\bibitem[{Blunsom and Cohn(2010)}]{blunsom2010unsupervised}
Phil Blunsom and Trevor Cohn. 2010.
\newblock Unsupervised induction of tree substitution grammars for dependency
  parsing.
\newblock In \emph{Proceedings of the 2010 Conference on Empirical Methods in
  Natural Language Processing}, pages 1204--1213. Association for Computational
  Linguistics.

\bibitem[{Chu and Liu(1965)}]{chu1965shortest}
Yoeng-Jin Chu and Tseng-Hong Liu. 1965.
\newblock On shortest arborescence of a directed graph.
\newblock \emph{Scientia Sinica}, 14(10):1396.

\bibitem[{Cohen et~al.(2008)Cohen, Gimpel, and Smith}]{cohen2008logistic}
Shay~B Cohen, Kevin Gimpel, and Noah~A Smith. 2008.
\newblock Logistic normal priors for unsupervised probabilistic grammar
  induction.
\newblock In \emph{Advances in Neural Information Processing Systems}, pages
  321--328.

\bibitem[{Edmonds(1967)}]{edmonds1967optimum}
Jack Edmonds. 1967.
\newblock Optimum branchings.
\newblock \emph{Journal of Research of the National Bureau of Standards B},
  71(4):233--240.

\bibitem[{Eisner(1996)}]{eisner1996three}
Jason~M Eisner. 1996.
\newblock Three new probabilistic models for dependency parsing: An
  exploration.
\newblock In \emph{Proceedings of the 16th conference on Computational
  linguistics-Volume 1}, pages 340--345. Association for Computational
  Linguistics.

\bibitem[{Gelling et~al.(2012)Gelling, Cohn, Blunsom, and
  Gra{\c{c}}a}]{gelling2012pascal}
Douwe Gelling, Trevor Cohn, Phil Blunsom, and Joao Gra{\c{c}}a. 2012.
\newblock The pascal challenge on grammar induction.
\newblock In \emph{Proceedings of the NAACL-HLT Workshop on the Induction of
  Linguistic Structure}, pages 64--80. Association for Computational
  Linguistics.

\bibitem[{Grave and Elhadad(2015)}]{grave2015convex}
Edouard Grave and No{\'e}mie Elhadad. 2015.
\newblock A convex and feature-rich discriminative approach to dependency
  grammar induction.
\newblock In \emph{ACL (1)}, pages 1375--1384.

\bibitem[{Jiang et~al.(2016)Jiang, Han, and Tu}]{jiang-han-tu:2016:EMNLP2016}
Yong Jiang, Wenjuan Han, and Kewei Tu. 2016.
\newblock \href {https://aclweb.org/anthology/D16-1073} {Unsupervised neural
  dependency parsing}.
\newblock In \emph{Proceedings of the 2016 Conference on Empirical Methods in
  Natural Language Processing}, pages 763--771, Austin, Texas. Association for
  Computational Linguistics.

\bibitem[{Klein and Manning(2004)}]{klein2004corpus}
Dan Klein and Christopher~D Manning. 2004.
\newblock Corpus-based induction of syntactic structure: Models of dependency
  and constituency.
\newblock In \emph{Proceedings of the 42nd Annual Meeting on Association for
  Computational Linguistics}, page 478. Association for Computational
  Linguistics.

\bibitem[{Koo et~al.(2007)Koo, Globerson, Carreras~P{\'e}rez, and
  Collins}]{koo2007structured}
Terry Koo, Amir Globerson, Xavier Carreras~P{\'e}rez, and Michael Collins.
  2007.
\newblock Structured prediction models via the matrix-tree theorem.
\newblock In \emph{Joint Conference on Empirical Methods in Natural Language
  Processing and Computational Natural Language Learning (EMNLP-CoNLL)}, pages
  141--150.

\bibitem[{Le and Zuidema(2015)}]{le2015unsupervised}
Phong Le and Willem Zuidema. 2015.
\newblock \href {http://www.aclweb.org/anthology/N15-1067} {Unsupervised
  dependency parsing: Let's use supervised parsers}.
\newblock In \emph{Proceedings of the 2015 Conference of the North American
  Chapter of the Association for Computational Linguistics: Human Language
  Technologies}, pages 651--661, Denver, Colorado. Association for
  Computational Linguistics.

\bibitem[{McDonald et~al.(2005)McDonald, Crammer, and
  Pereira}]{mcdonald2005online}
Ryan McDonald, Koby Crammer, and Fernando Pereira. 2005.
\newblock Online large-margin training of dependency parsers.
\newblock In \emph{Proceedings of the 43rd annual meeting on association for
  computational linguistics}, pages 91--98. Association for Computational
  Linguistics.

\bibitem[{Naseem et~al.(2010)Naseem, Chen, Barzilay, and
  Johnson}]{naseem2010using}
Tahira Naseem, Harr Chen, Regina Barzilay, and Mark Johnson. 2010.
\newblock Using universal linguistic knowledge to guide grammar induction.
\newblock In \emph{Proceedings of the 2010 Conference on Empirical Methods in
  Natural Language Processing}, pages 1234--1244. Association for Computational
  Linguistics.

\bibitem[{Paskin(2001)}]{paskin2001cubic}
Mark~A Paskin. 2001.
\newblock \emph{Cubic-time parsing and learning algorithms for grammatical
  bigram models}.
\newblock Citeseer.

\bibitem[{Spitkovsky et~al.(2013)Spitkovsky, Alshawi, and
  Jurafsky}]{spitkovsky2013breaking}
Valentin~I Spitkovsky, Hiyan Alshawi, and Daniel Jurafsky. 2013.
\newblock Breaking out of local optima with count transforms and model
  recombination: A study in grammar induction.
\newblock In \emph{EMNLP}, pages 1983--1995.

\bibitem[{Spitkovsky et~al.(2010)Spitkovsky, Alshawi, Jurafsky, and
  Manning}]{spitkovsky2010viterbi}
Valentin~I Spitkovsky, Hiyan Alshawi, Daniel Jurafsky, and Christopher~D
  Manning. 2010.
\newblock Viterbi training improves unsupervised dependency parsing.
\newblock In \emph{Proceedings of the Fourteenth Conference on Computational
  Natural Language Learning}, pages 9--17. Association for Computational
  Linguistics.

\bibitem[{Tarjan(1977)}]{tarjan1977finding}
Robert~Endre Tarjan. 1977.
\newblock Finding optimum branchings.
\newblock \emph{Networks}, 7(1):25--35.

\bibitem[{Tu and Honavar(2012)}]{tu2012unambiguity}
Kewei Tu and Vasant Honavar. 2012.
\newblock Unambiguity regularization for unsupervised learning of probabilistic
  grammars.
\newblock In \emph{Proceedings of the 2012 Joint Conference on Empirical
  Methods in Natural Language Processing and Computational Natural Language
  Learning}, pages 1324--1334. Association for Computational Linguistics.

\end{thebibliography}
\bibliographystyle{emnlp_natbib}

\end{document}